# Enhancing Crash Frequency Modeling Based on Augmented Multi-Type Data by Hybrid VAE-Diffusion-Based Generative Neural Networks

Junlan Chen[1,2], Qijie He[1,2], Pei Liu[3], Wei Ma[4], Ziyuan Pu[1,5], Nan Zheng[1,2]

[1]School of Transportation, Southeast University, No.2 Southeast University Road, Nanjing, China, 211189. E-mail: junlan.chen@monash.edu; qhee0026@student.monash.edu; ziyuan.pu@monash.edu

[2]School of Civil Engineering, Monash University, Melbourne, Australia. E-mail: junlan.chen@monash.edu; nan.zheng@monash.edu

[3]Intelligent Transportation Thrust, Systems Hub, The Hong Kong University of Science and Technology (Guangzhou), Guangzhou, China. E-mail: pliu061@connect.hkust-gz.edu.cn

[4]Department of Civil and Environmental Engineering, The Hong Kong Polytechnic University, Hung Hom, Hong Kong. E-mail: wei.w.ma@polyu.edu.hk

[5]School of Engineering, Monash University, Jalan Lagoon Selatan, 47500 Bandar Sunway, Malaysia.

**ABSTRACT**

Crash frequency modeling is used to analyze how explanatory factors, such as traffic volume, road geometry, and environmental conditions, influence the frequency of crashes. Inaccurate predictions can distort the understanding of how different explanatory factors contribute to crashes. As a result, they can lead to misinformed policy decisions and wasted resources, ultimately jeopardizing traffic safety. Nevertheless, crash frequency modeling is often constrained by excessive zero observations in crash datasets, a limitation rooted in traditional data collection methods that rely on historical crash records, such as police reports, insurance claims, and hospital records. The low probability occurrence of crashes, unreported minor crashes, and the high costs associated with the data collection process collectively lead to a high prevalence of zero crash count. This problem can reduce the accuracy of crash frequency modeling, leading to biased predictions that impede safety policy-making. Existing approaches to address excessive zero observations primarily include statistical methods, data aggregation and resampling techniques. While statistical methods often rely on restrictive distributional assumptions, data aggregation and sampling techniques can lead to significant information loss and introduce biases that distort the representation of crash data. To overcome these limitations and provide a more cost-effective alternative for crash data collection, we introduce deep generative models that can augment crash data by producing high-quality synthetic samples and reducing the excess zero observations. However, while existing deep generative models have made progress in synthetic tabular data generation, they often struggle with the complexity of multi-type tabular crash data, which includes count, ordinal, nominal, and real-valued variables. In addition, typical deep generative models rely on simplistic one-hot encoding techniques, leading to dimensional expansion, which increases computational complexity and hinders models' ability to capture correlations between features, ultimately limiting their ability to generate realistic synthetic crash data. Therefore, this study proposes a hybrid VAE-Diffusion neural network to address excessive zero observations in crash data and tackle the data generation challenges posed by the complexities of multi-type tabular crash data. We evaluate the quality of the synthetic data generated by the proposed model against baseline models using multiple metrics such as similarity, accuracy, diversity, and structural consistency. Furthermore, we compare the predictive performance of a machine learning model trained on augmented crash data with that of a traditional statistical model. Finally, we interpret the crash frequency prediction results and provide policy recommendations to enhance traffic safety. By effectively capturing the complex patterns in multi-type crash data, the proposed hybrid VAE-Diffusion model surpasses baseline models across all metrics in terms of synthetic data quality and predictive accuracy. The findings of this study highlight the potential of synthetic data to improve crash frequency modeling and provide actionable recommendations to enhance traffic safety.

# 1 INTRODUCTION

Crashes are complex and multidimensional events where many interrelated factors can contribute to their occurrence. To account for the complexity and randomness of crash data, crash frequency modeling is widely employed as a useful tool to analyze how various explanatory factors, e.g., traffic volume, road geometry, and environmental conditions, correlate to the aggregated number of crashes that occurred in a given time at different locations (Lord et al., 2021). Accurate crash frequency modeling is essential in assisting governmental agencies to develop effective safety policies and allocate resources to high-risk areas, ultimately improving traffic safety and reducing crashes (Lord and Mannering, 2010). However, the crash data are often characterized by excess zero observations, indicating that a significant portion of the data consists of zero crash counts (Ali et al., 2024; Dzinyela et al., 2024). This can be problematic in crash frequency modeling because excess zeros can bias parameter estimates, resulting in erroneous predictions and misleading inferences about explanatory factors (Gu et al., 2020; Raihan et al., 2019).

The root of having excess zero observations in crash data lies in the limitations of the traditional data collection method. Crash data are often collected from police reports, insurance claims, and hospital records after crashes occurred and the data collection process is typically costly (Lord et al., 2021). Given that crashes are low-probability events, post-crash data collection methods naturally lead to a small sample size of recorded crashes. Moreover, many minor crashes are not documented due to factors such as perceived insignificance or absence of police involvement, which further reduces the sample size of recorded crashes (Lord et al., 2005). These limitations collectively result in a crash count distribution that skews heavily toward zero. Although there are emerging technologies, such as data collection from on-road electronic equipment like video surveillance, radars, loop detectors (Maha Vishnu et al., 2017), and from onboard equipment installed on vehicles to gather crash data related to vehicle conditions, driver distractions, traffic information, etc (Josephinshermila et al., 2023). However, since crashes are rare, such collection methods still struggle with the problem of excess zero observations, as they cannot capture data where crashes are nonexistent.

Approaches to address excessive zero observations in crash data include both traditional model-driven statistical models and data-centric techniques. Model-driven statistical models, such as zero-inflated Poisson and zero-inflated negative binomial (Lee and Mannering, 2002) are often constrained by assumptions of data distribution and may not fully capture the underlying patterns in crash datasets with multiple data types (i.e., count, ordinal, nominal, and real-valued) and complex interactions between various explanatory variables (Ding et al., 2022a). Alternatively, data-centric techniques provide greater flexibility to handle excessive zero observations. An approach is to aggregate crash data over extended time periods or across longer road segments, though this can result in information loss and reduce reliability in crash frequency prediction (Usman et al., 2011). Resampling techniques have also been used, including under-sampling and over-sampling (Pei et al., 2016). However, under-sampling can discard potentially valuable data, while over-sampling methods suffer from generating minority samples that are most likely to be majority, which can be deceptive and lead to severe failures in real-world applications (Tarawneh et al., 2022).

Deep generative models are effective techniques in alleviating excessive zero observations; meanwhile, they can reduce the time-consuming and expensive crash data collection process (Missaoui et al., 2023). Deep generative models show strong performance on image data with continuous pixel values and local spatial correlations, yet face challenges with heterogeneous tabular data due to its multi-type features and lack of inherent spatial relationships (Borisov et al., 2024a; Kim et al., 2022). Several deep generative models show improvement in tabular data generation, such as Conditional Tabular Generative Adversarial Networks (CTGAN)(Xu et al., 2019), tabular variational autoencoder (TVAE) (Xu et al., 2019), and Generation of Realistic Tabular data (GReaT)(Stoian et al., 2024). Recently, a novel Diffusion model has demonstrated even greater performance in generating data with high quality and diversity (Kim et al., 2023). Despite its impressive capabilities, applying diffusion models to tabular data still presents challenges (Kotelnikov et al., 2022). One of the key challenges is that encoding techniques commonly used in deep generative models face difficulties in handling tabular crash data (Borisov et al., 2024b). Most existing models adopted simple encoding techniques such as one-hot encoding (Lee et al., 2023; Liu et al., 2023) and analog bit encoding (Zheng and Charoenphakdee, 2022) to convert discrete features into numerical format. However, the crash dataset is highly structured multi-type data with high-cardinality variables such as count and ordinal variables, which simple encoding methods struggle to handle. Simple encoding may aggravate the 'curse of dimensionality' problem, expanding tabular crash data into high-dimensional vectors. This dimensional expansion in the data increases computational complexity as the models struggle to capture meaningful correlations across features. Consequently, this undermines the model's ability to generate realistic synthetic data (Lee et al., 2023; Liu et al., 2023).

To this end, this study introduces a data-driven crash frequency modeling approach based on the advanced VAE-Diffusion deep generative model (Zhang et al., 2023). Following data generation, we compare the proposed model with the baseline generative models, in terms of the synthetic data quality and prediction accuracy. Lastly, we interpret the prediction result and provide policy recommendations for traffic safety and management.

The main contributions of the study are summarized below:

(1) A novel deep generative neural network designed to process multi-type crash data by transforming diverse input features into a unified embedding space, where the model can simultaneously learn and extract various features. It generates synthetic data that mirrors real-world crash data, thus resolving excessive zero observations and offering a more efficient and cost-effective alternative for data collection.

(2) The hybrid VAE-Diffusion architecture of the generative model, integrating a VAE with a Transformer encoder and decoder, maps all features to a lower-dimensional space, mitigating the 'curse of dimensionality.'

(3) A comprehensive study was conducted to evaluate the quality of the synthetic data with complex multi-types features, using coverage similarity metrics like distribution plots and statistical metrics, along with structural similarity metrics like Pair-wise Correlation Difference (PCD).

## 2 LITERATURE REVIEW

2.1 Crash Frequency Modelling

Crash frequency modeling serves as a vital tool for analyzing nonnegative and discrete crash events on roadways, enabling the assessment of safety performance across entities such as roadway segments, intersections, and corridors(Lord et al., 2021). Conventional statistical models for handling excessive zero observations often adapt count-data models to better accommodate the large number of zero observations. Raihan et al. (2019) and Lord et al. (2005) applied zero inflation Poisson model and the zero-inflation negative binomial model, which assume two processes: one where the roadway section is virtually safe with zero crashes, and another where the section is in a non-negative count state for crashes, following a Poisson or negative binomial distribution. Yet, according to Lord et al. (2007), these models have intrinsic logical shortcomings, as they fail to establish clear boundary conditions to differentiate fully safe states from unsafe states. Subsequently, the negative binomial Lindley model, a mixture of negative binomial and Lindley distributions (Ghitany et al., 2008; Lindley, 1958), was introduced to address excess zero observations (Behara et al., 2021; Rusli et al., 2018; Shinthia Azmeri et al., 2023). Building on this, Islam et al. (2022) employ a Finite Mixture Negative Binomial-Lindley (FMNB-L) model, combining the Negative Binomial-Lindley distribution with finite mixture modeling to account for heterogeneous subpopulations and excess zero observations in crash data. However, similar to other statistical models, a notable limitation of this approach is that the increased complexity of the model restricts its applicability to relatively small datasets. Owing to the limitations of statistical models, such as predefined assumptions and inability to handle large datasets, machine learning-based approaches have gained traction for addressing excessive zero observations.

Compared to statistical models,machine learningg modelsoutperformm in predictionass they are flexible, require no prior assumptions about data distribution, and can learn complex patterns in high-dimensional data (Li et al., 2008; Zeng et al., 2016a). For example, Chang and Chen (2005) employed the CART (Classification and Regression Trees) method, a decision tree-based approach that recursively splits data into subsets to model crash frequency. They found that the model achieved better prediction accuracy compared to a negative binomial regression model, with accuracy rates of 52.6% and 52.3%, respectively. Das and Abdel-Aty (2011) model crash frequency using a Genetic Programming (GP) approach and conclude that GP is effective in capturing the relationship between crash frequency and explanatory factors such as average daily traffic, roadway conditions, and skid resistance, providing a comprehensive understanding of rear-end crashes on urban arterials. Zeng et al. (2016) employs an optimized Neural Network (NN) approach and conclude that the structure-optimized NN model not only outperforms traditional

Negative Binomial (NB) models in terms of fitting and predictive performance but also effectively identifies and removes variables with insignificant effects on crash frequency, resulting in a more interpretable and simplified model. Additionally, the extracted rules confirm the existence of nonlinear relationships between crash frequency and explanatory factors. Although machine learning models excel at capturing complex nonlinear relationships in crash frequency modeling, their performance is still limited by data quality, including the completeness and representativeness of the available data (Mannering et al., 2020).

2.2 Resampling Techniques for Multi-Type Crash Frequency Data

Previous studies focusing on data-centric approaches often utilize data aggregation or resampling techniques to solve excessive zero observations. Data aggregation involves adjusting space and time scales to reduce the number of zero observations while maintaining the homogeneity of the observations (e.g., lane width, shoulder width, etc.). For example, aggregate 200-meter road segments into 400-meter segments, combined with aggregation over different time scales (e.g., 2, 3, 4, and 6 years) to decrease the proportion of zero observations before developing regression models. However, this may lead to a loss of spatial-temporal information and compromise the accuracy of crash frequency prediction (Usman et al., 2011). Traditional resampling techniques are more sophisticated methods to address excessive zero observations. Abdel-Aty et al. (2004) and Yang et al. (2018) use matched case-control design to select a matched sub-set out of the full dataset, controlling for external factors such as time of day, season, year, geometric and roadway features, etc. Yet such under-sampling methods may lead to a loss of important information and potentially degrade model performance (Cai et al., 2020). On the other hand, Chen et al. (2022) applied the Synthetic Minority Over-sampling Technique for panel data (SMOTE-P), which increases minority class samples to balance the data, to deal with excessive zero observations of bus-involved crashes and improve model accuracy. Nevertheless, since SMOTE uses interpolation of existing samples, it can oversimplify relationships between variables, generating synthetic data that fails to capture the full complexity of crash patterns or reflect real-world variability, potentially leading to overfitting or biased predictions (Branco et al., 2016; Tarawneh et al., 2022). This limitation has motivated the adoption of deep generative models that can better represent and address the complexities inherent in crash frequency data (Chen et al., 2024).

Crash frequency data often contain multi-type variables (i.e., count, ordinal, nominal, and real-valued variables), posing significant challenges for deep generative models in addressing excessive zero observations. Most existing studies use Generative Adversarial Networks (GANs) (Cai et al., 2020; Man et al., 2022) or variational autoencoder (VAE) (Ding et al., 2022b) to address class imbalance in crash data. As deep generative models evolve, various models built on different foundational architectures emerge (Zhang et al., 2023). Although these methods offer advantages for multi-type data synthesis, concerns remain about data quality due to their limitations in capturing the complex relationships between variables. Conditional Tabular Generative Adversarial Networks (CTGAN) and tabular variational autoencoder (TVAE) (Xu et al., 2019) both utilize the Mode-specific Normalization model and treat each category of discrete variables as a separate mode and normalize the continuous variables within each mode. However, this method can introduce model complexity that can obscure the features of the variables and may fail to capture potential relationships between different categories. Score-based Tabular Data Synthesis (StaSy) (Kim et al., 2023) is a diffusion-based model, which employs a technique that treats one-hot encoded categorical variables as continuous features and processes them together with numerical columns. Nevertheless, treating all variables as continuous may result in high-dimensional feature spaces and make it difficult for the model to learn data features and capture intrinsic features of the discrete variables. Generation of Realistic Tabular data (GReaT) (Borisov et al., 2022) employs a serialization technique that converts both discrete and continuous variables into natural language sentences. Then the 'sentences' can be processed by an auto-regressive GPT model. However, the natural language descriptions can lead to loss of detail or ambiguity, as similar values might be described in similar ways, affecting the model's ability to capture precise data features. Additionally, while shuffling rows ensures permutation invariance, it may not fully capture the relationships and structures between columns, particularly when column order is important for analysis.

## 3 METHODS

### 3.1 Multi-Type Data Generative Neural Network

We propose a Hybrid VAE-Diffusion deep generative model to generate synthetic samples of the minority class (i.e., non-zero crash samples) to rebalance the data. The model is comprised of two main components: a Variational Autoencoder (VAE) and a Diffusion data generator, as shown in Fig.2. The VAE, with a Transformer encoder and decoder, turns the data into a machine-readable form, so that the diffusion model can better understand and learn the intra- and inter-column patterns of the dataset. With the proposed two-component structure, the model can better handle and learn from the four-multi-type, highly-structured tabular crash data, and generate synthetic data of outstanding quality.

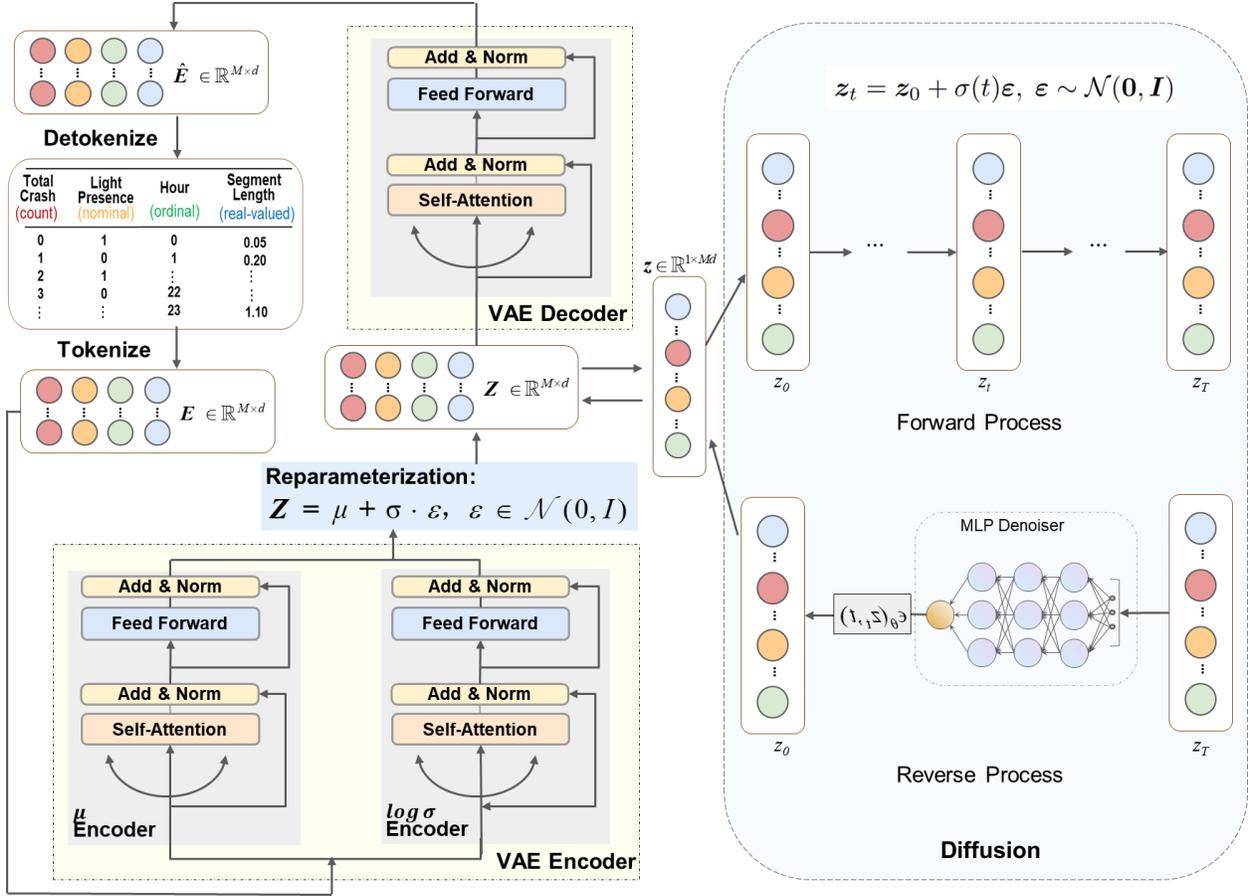

Fig. 2 The Hybrid VAE-Diffusion Multi-Type Data Generation Model Architecture for Enhancing Crash Frequency Prediction

3.2 Problem Definition

Each column in the crash data table is a variable and each row is a sample. In this study, the crash dataset contains four types of variables (or columns): ordinal, nominal, count, and real-valued. The first three are discrete variables: nominal variables represent categorical data without any inherent order (e.g., the presence of light: 0 = no, 1 = yes); ordinal variables have a natural order (e.g., the hour of the day: 0, 1, 2, ..., 23); and count variables indicate the frequency of occurrences using non-negative integers (e.g., crash frequency: 0, 1, 2, 3, ...). Real-valued variables are continuous variables that can take any value within a range, such as the length of the road segments.

$M_{dis}$ and $M_{con}$ denote the number of discrete and continuous columns (or variables), respectively. Each row in the crash data table, corresponding to the record of the crash frequency that occurred in a certain road segment in the specific hour of the year along with the road information, is represented as a vector $x = [x^{dis}, x^{con}]$, where $x$ has a dimensionality of ($M_{dis} + M_{con}$). The i-th discrete variable $x_i^{dis}$ has $C_i$ candidate values, where $x_i^{dis} \in \{1, ..., C_i\}$. We input the original training dataset $\mathcal{T} = \{x\}$ to the model to learn the parameterized generative model $p_\theta(\mathcal{T})$, where synthetic data $\hat{x} \in \hat{\mathcal{T}}$ can be generated.

3.3 Data Transformation with VAE-Transformer Model

Data transformation refers to the process of converting the tabular data into a model-learnable format—column embedding. Effective transformation enables the model to extract meaningful information signal from the data (Borisov et al., 2024a). The transformation process involves several steps: Feature Tokenization, Transformer Encoding and Decoding, and Detokenization.

Feature Tokenization. Tokenization is the first step is to map all the columns into a d-dimensional vector—the initial column embedding. We apply one-hot encoding to preprocess discrete variables, i.e., count, ordinal, and nominal variables, denote as $x_i^{one-hot} \in \mathbb{i}^{1 \times C_i}$. Then each row (or record) can be represented as $x = [x^{con}, x_1^{one-hot}, \mathrm{L}, x_{M_{dis}}^{one-hot}] \in \mathbb{?}^{M_{con} + \sum_{i=1}^{M_{dis}} C_i}$, where x is a ($M_{con} + \sum_{i=1}^{M_{dis}} C_i$) dimension vector. Then create a d-dimensional vector for continuous and discrete columns, by applying linear transformation and create an embedding lookup table for the two types of columns, respectively.

$$e_i^{con} = x_i^{con} \cdot w_i^{con} + b_i^{num}, \quad w_i^{con}, b_i^{num}, e_i^{con} \in \mathbb{R}^{1 \times d}, \quad (1)$$

$$e_i^{dis} = x_i^{one-hot} \cdot W_i^{dis} + b_i^{dis}, \quad b_i^{dis}, e_i^{dis} \in \mathbb{R}^{1 \times d}, W_i^{dis} \in \mathbb{R}^{C_i \times d}, \quad (2)$$

where $w_i^{con}, b_i^{num}, b_i^{dis}, W_i^{dis}$ are learnable parameters of the tokenizer. Stack all the rows into the embeddings of all column, denoted as E, which is the initial column embedding:

$$E = [e_1^{con}, \ldots, e_{M_{con}}^{con}, e_1^{dis}, \ldots, e_{M_{dis}}^{dis}] \in \mathbb{R}^{M \times d}. \quad (3)$$

Transformer Encoding and Decoding. Simple tokenization might lead to the model's incomplete understanding of each variable's unique characteristics, especially in the crash dataset like ours that includes ordinal, nominal, count, and real-valued variables. We introduce a Transformer encoder-decoder to perform the multi-dimensional encoding, which effectively aids our model to understand the variables features and complexity of the data pattern.

The encoder and decoder are both two-layer transformers, each with a Self-Attention module and a Feed Forward Neural Network (FFNN) denoted as H0, which is a simple two-layer MLP. The loss function for constructing the VAE is in Equation (4).

$$\begin{cases} H_1 = \text{Re}LU(FC(H_0)) \in \mathbb{R}^{M \times D}, \\ H_2 = FC(H_1) \in \mathbb{R}^{M \times d}, \end{cases} \quad (4)$$

$$\mathcal{L} = l_{recon}(x, \hat{x}) + \beta l_{kl} \quad (5)$$

The VAE encoders are designed to generate $\mu$ and $\sigma$ to construct the latent embedding $Z$, shown as Equation (6):

$$Z = \mu + \sigma \cdot \varepsilon, \quad \varepsilon \in \mathcal{N}(0, I) \quad (6)$$

Detokenization. Lastly, we detokenize and reconstruct token representation of each column to its original human-readable format. The process of detokenization is symmetric to that of tokenization:

$$\hat{x}_i^{con} = \hat{e}_i^{con} \cdot \hat{w}_i^{con} + \hat{b}_i^{con}, \quad \hat{x}_i^{one-hot} = Softmax(\hat{e}_i^{dis} \cdot \hat{W}_i^{dis} + \hat{b}_i^{dis}), \quad (7)$$

$$\hat{x} = [\hat{x}_1^{dis}, \ldots, \hat{x}_{M_{dis}}^{dis}, \hat{x}_1^{one-hot}, \ldots, \hat{x}_{M_{cat}}^{one-hot}], \quad (8)$$

where $\hat{w}_i^{con} \in \mathbb{R}^{d \times 1}$, $\hat{b}_i^{con} \in \mathbb{R}^{1 \times 1}$, $\hat{W}_i^{dis} \in \mathbb{R}^{d \times C_i}$, $\hat{b}_i^{dis} \in \mathbb{R}^{1 \times C_i}$ are detokenizer's parameters.

3.4 Diffusion Data Synthesis

The Diffusion model is designed to learn the underlying distribution of crash data and generate synthetic data. Inspired by the thermodynamic process where particles spread out until evenly distributed, the model adds and spreads out noises to the data. It then learns to reverse this process, removing the noise step-by-step to create clear, structured samples. Instead of directly learning the distribution of the data itself, the Diffusion model learns how to denoise the data. This strategy allows it to generate diverse data and reduces the likelihood of mode collapse.

After the VAE-Transformer model is well-learned, latent embedding $Z$ can be extracted from the encoder, flattened, and then transfer into a vector as $z = Flatten(Encoder(x)) \in \mathbb{R}^{1 \times Md}$. The distribution of the latent embeddings is $p(z)$, learned by the diffusion model via two processes—a forward adding noise process and a reverse denoising process. In the forward process, Stochastic Differential Equation (SDE) and Variance Exploding (VE) (Song et al., 2020) are applied to acquire the perturbed data $z_t$, which is the diffused embedding at time t, and $\sigma(t)$ is the noise level, shown as Equation (10). In the reverse process, $\nabla_{z_t} \log p(z_t) dt$ is the score function of $z_t$, $\omega_t$ is the standard Wiener process. The loss function is in Equation (11) (Karras et al., 2022), where $\epsilon_\theta(z_t,t)$ is the denoising function, which is used to estimate the Gaussian noise.

$$z_t = z_0 + \sigma(t)\varepsilon, \quad \varepsilon \sim \mathcal{N}(0,I) \quad \text{(Forward Process)} \quad (9)$$

$$dz_t = -2\dot{\sigma}(t)\sigma(t)\nabla_{z_t} \log p(z_t)dt + \sqrt{2\dot{\sigma}(t)\sigma(t)}\,d\omega_t \quad \text{(Reverse Process)} \quad (10)$$

$$\mathcal{L} = \mathbb{E}_{z_0 \sim p(z_0)}\mathbb{E}_{t \sim p(t)}\mathbb{E}_{\varepsilon \sim \mathcal{N}(0,I)} \|\epsilon_\theta(z_t,t) - \varepsilon\|_2^2, \text{ where } z_t = z_0 + \sigma(t)\varepsilon \quad \text{(Loss Function)} \quad (11)$$

To obtain the denoising function $\epsilon_\theta(z_t,t)$, a five-layer Multilayer Perceptron (MLP) is trained, where the hidden layer is fully connected(Kotelnikov et al., 2022), as shown in Equation (12). Lastly, the denoising function can be applied to the loss function for model training in Equation (11).

$$\begin{cases} h_0 = FC_{in}(z_t) \in \mathbb{R}^{1 \times d_{hidden}} \\ h_{in} = h_0 \oplus t_{emb} \\ h_1 = SiLU(FC_1(h_0) \in \mathbb{R}^{1 \times 2*d_{hidden}}) \\ h_2 = SiLU(FC_2(h_1) \in \mathbb{R}^{1 \times 2*d_{hidden}}) \\ h_3 = SiLU(FC_3(h_2) \in \mathbb{R}^{1 \times d_{hidden}}) \\ \epsilon_\theta(z_t,t) = h_{out} = FC_{out}(h_3) \in \mathbb{R}^{1 \times d_{in}} \end{cases} \quad (12)$$

**4 Experimental Design**

4.1 Generation and Evaluation Framework

This study aims to address the excessive zero observations and enhance the overall performance of crash frequency modeling. The data generation and evaluation framework is shown in based on its coverage and structural similarity to the real data. We adopt various metrics, including distribution plots, Detection Score (C2ST), α-Precision and β-Recall. While the structural similarity is appraised by Pair-wise Correlation Difference (PCD).

Fig. 1. We compare the quality of synthetic data generated by both the proposed multi-type data generation model, and the baseline models. The quality of synthetic data can be assessed

We use a machine learning (ML) model, XGBoost, for prediction. To evaluate the modeling accuracy of each resampling method, we compared crash frequency prediction models that were fitted using data balanced by different generative models. The relevant metrics, calculated using the test set, include RMSE and MSE. We also compare the accuracy of the proposed ML-based model to classic statistical models—Zero-inflated Poisson Regression. Since ML-based models can often be perceived as black boxes due to their complexity, we use SHAP (SHapley Additive exPlanations) to visualize and provide an intuitive explanation of the prediction results.

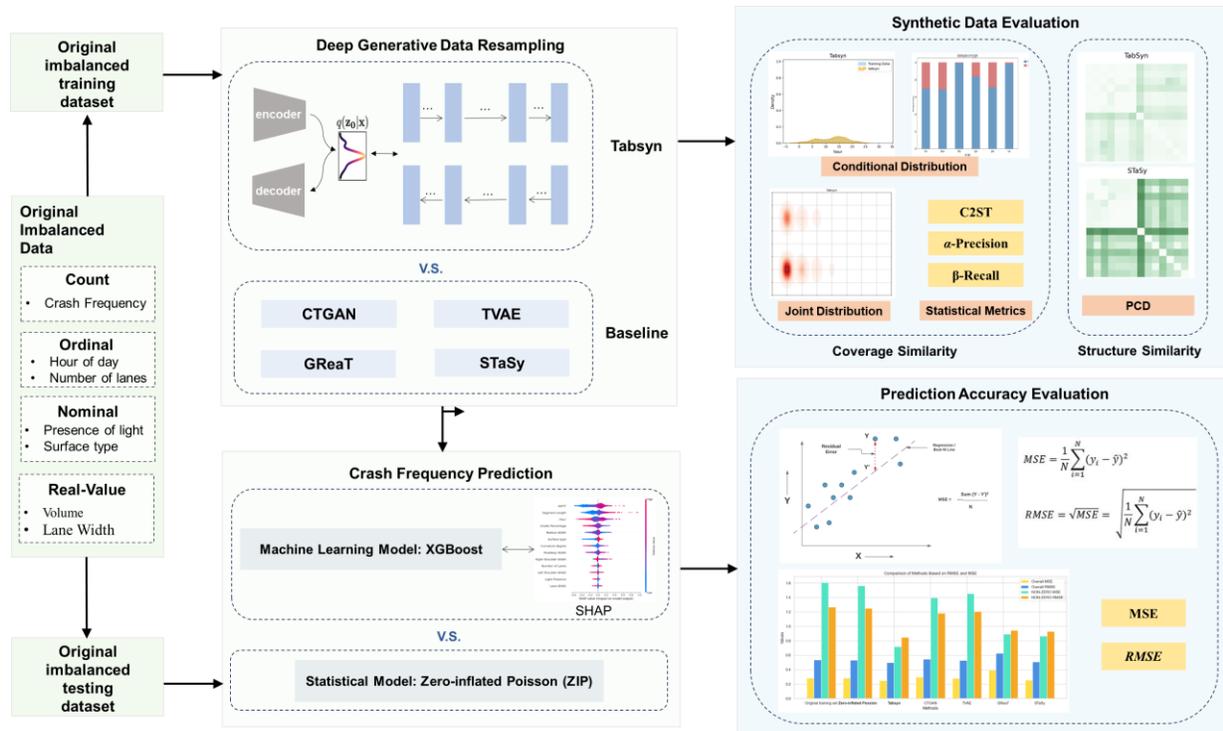

Fig. 1 The framework of data generation and the designed evaluation experiments.

4.2 Data Description

This study focuses on crashes on the mainline of Interstate 5 (I-5) from milepost 139.69 to 210.10 in Washington State in 2019. The crash dataset, sourced from the Highway Safety Information System (HSIS) (Anusha patel Nujjetty, 2022), includes roadway and crash files. The roadway file provides information for each road segment, such as the number of lanes, presence of lighting, lane width, segment length, and other details. The crash file contains officer-reported crash incidents.

We select road segments longer than 0.05 miles (approximately 64 meters) () to reduce the potential data coding errors. To learn the impact of time factors on crash frequency, we count the crashes that happened in each segment for every hour of the day throughout the year. The number of crashes on each segment for each hour within a 24-hour period, along with the segment's geometric information, is treated as a record. As a result, there are a total of 3714 crashes across 372 road segments in both directions in 2019. The total number of records is 372 (road segments) * 2 (directions) * 24 (hours) * 1(year) = 17856, of which 15142 (84.8%) have zero observations. The prevalence of zero observations in our dataset indicates a significant excess zero problem. The summary statistics of variables included in this research are shown in Table 1.

Table 1 Data Description

| Variables | Description | Non-zero Records | | | | Zero Records | | | |
|---|---|---|---|---|---|---|---|---|---|
| | | Mean | STD | Min | Max | Mean | STD | Min | Max |
| **Dependent Variable** | | | | | | | | | |
| *Count* | | | | | | | | | |
| Total_Crash | crash occurrence per segment per year per hour | 1.368 | 0.796 | 1 | 10 | 0 | 0 | 0 | 0 |
| | | | | | | | | | |
| **Explanatory Variables** | | | | | | | | | |
| *Ordinal* | | | | | | | | | |
| Hour | The hour when crashes happened | 12.443 | 5.624 | 0 | 23 | 11.331 | 7.117 | 0 | 23 |

| Variable | Description | | | | | | | |
|---|---|---|---|---|---|---|---|---|
| Number_of_Lanes | Number of lanes (in one direction) | 3.552 | 0.556 | 2 | 5 | 3.484 | 0.583 | 2 | 5 |
| *Nominal* | | | | | | | | | |
| Light_Presence | Presence of light (0 for absence;1 for presence) | 0.299 | - | 0 | 1 | 0.306 | - | 0 | 1 |
| Surface Type | Surface type (0 for Asphalt; 1 for Concrete) | 0.671 | - | 0 | 1 | 0.595 | - | 0 | 1 |
| *Real-Value* | | | | | | | | | |
| AAHT | Annual average hourly traffic flow | 3.570 | 1.775 | 0.0001 | 8.620 | 2.661 | 1.751 | 0.0001 | 8.6199 |
| Segment_Length | Segment Length (mile) | 0.188 | 0.189 | 0.050 | 1.100 | 0.134 | 0.124 | 0.050 | 1.100 |
| Lane_Width | Lane Width (ft) | 12.467 | 1.483 | 11 | 22 | 12.355 | 1.329 | 11 | 22 |
| Roadway_Width | Total roadway width for the roadway segment (ft) | 87.804 | 14.760 | 55 | 132 | 85.464 | 14.443 | 55 | 132 |
| Median_Width | Median width of the segment | 56.083 | 50.650 | 7 | 300 | 60.482 | 51.853 | 7 | 300 |
| Left_Shoulder_Width | Left shoulder width of the segment (ft) | 8.530 | 3.496 | 0 | 12 | 8.970 | 3.013 | 0 | 12 |
| Right_Shoulder_Width | Right shoulder width of the segment (ft) | 8.530 | 3.496 | 0 | 12 | 8.970 | 3.013 | 0 | 12 |
| Curvature_degree | Curvature degree | 0.747 | 0.969 | 0 | 4 | 0.672 | 0.997 | 0 | 4.070 |
| Grade_Percentage | Grade percentage | -0.055 | 2.250 | -5 | 5 | 0.123 | 2.086 | -5 | 5 |

Note: * Hour, hour of the year when crashes occurred: 0 for 12 p.m.; 1 for 1 a.m.; 2 for 2 a.m.; …; 22 for 10 p.m.; 23 for 11 p.m.

* Grade Percentage: +for upgrade; - for downgrade

## 4.3 Data Resampling

To address excessive zero observations in our dataset, we use the deep generative over-sampling technique (i.e., VAE-Difussion, STaSy, CTGAN, GReaT, and TVAE) to resample our imbalanced crash dataset. We take the training set as the input for the generative models. They then can learn the data distribution and generate synthetic non-zero crash records for rebalancing.

The test set is for evaluating the synthetic data quality and prediction accuracy, which should remain the same during the whole modeling process.

Table 2 provides a summary of how the data is organized for model training and testing. The original dataset was randomly split into a 70% training set (with 12498 records) and a 30% testing set (with 5358 records). In this study, we adopt the balanced ratio of 1:1 (zero-crash records to non-zero records). Therefore, 12428 non-zero crash records should be generated and combined with the original total data set to get the balanced dataset.

Table 2 Experiment Design

| Data type | Original total dataset | Original test dataset | Original training dataset | Balanced total dataset |
|---|---|---|---|---|
| Zero-crash records | 15142 | 4543 | 10599 | 15142 |
| Non-zero records | 2714 | 815 | 1899 | 2714+**12428**=15142 |
| Total | 17856 | 5358 | 12498 | 30284 |

4.4 Model Performance Evaluation

To validate the effectiveness of the proposed approach, we structured the experiment in two parts. First, we evaluated the quality of the synthetic data generated by different generative models. Second, we compared the prediction accuracy of these data-driven modeling approaches with a statistical model.

*4.4.1 Baseline Models for Comparison*

We compare four baseline deep generative models against our proposed multi-type data generation model, including Conditional Tabular Generative Adversarial Networks (CTGAN), tabular variational autoencoder (TVAE), Generation of Realistic Tabular data (GReaT), and Score-based Tabular Data Synthesis (StaSy). These baseline models are chosen as they represent key generative approaches based on Generative Adversarial Networks (GANs), Variational Autoencoders (VAEs), and Diffusion models, making them relevant for comparison with our proposed model. Each of these models learns the underlying data distribution differently. CTGAN and VAE are based on two fundamental generative models, namely GANs and VAEs. GReaT, inspired by the transformer-based large language models (LLMs), treats each row of tabular data as a sentence and learns sentence-level distributions. And StaSy is a diffusion-based generative model. We evaluate the quality of the synthetic data they generate. The evaluation focuses on coverage and structural similarity. We then use the data generated by these models to predict crash frequency with XGBoost.

We also compared our data-driven approach with the traditional statistical model, the Zero-inflated Poisson (ZIP) model. The ZIP model is designed to handle count data with excess zeros. It assumes that two separate processes generate the data: A process that generates only zeros (zero-accident state); and another Poisson process that generates non-negative counts, including zeros(Lee and Mannering, 2002). The ZIP model can be expressed as:

$$P(n_{ij}=0) = p_i + (1-p_i)e^{-\lambda ij} \quad (13)$$

$$P(n_{ij}=n) = (1-p_i)\frac{\lambda_{ij}^n e^{-\lambda ij}}{n!}, n>0 \quad (14)$$

where $n_{ij}$ is the number of accidents on the roadway section $i$ during period $j$; $p_{ij}$ is the probability that segment $i$ is in the zero-accident state; and $\lambda_{ij}$ is the expected number of accidents for segment $i$ in period $j$ when it is not in the zero-accident state.

*4.4.2 Synthetic Data Quality Metrics*

The quality of synthetic data is evaluated by how similar its coverage and structure are to the real data. The coverage similarity was evaluated via distribution plots and statistical metrics like Detection Score (C2ST), $\alpha$-Precision and $\beta$-Recall. While the structural similarity is appraised by Pair-wise Correlation Difference (PCD).

**Detection Score (C2ST).** Classifier Two Sample Test (C2ST) is a metric that evaluates how difficult it is to tell apart the real data from the synthetic data. To calculate C2ST, first, create a table combining all rows of real and synthetic data, have all rows labeled, split it into training and validation sets, and then train a machine learning model—Logistic Regression—to predict these labels. The score is derived from the model's performance, typically using based on the average ROC AUC score across all the cross-validation splits. The C2ST score ranges from 0 to 1, where values closer to 1 indicate greater similarity to the real data, given by:

$$score = 1 - (max(ROC\ AUC, 0.5) \times 2 - 1) \quad (15)$$

**α-Precision** and **β-Recall.** α-Precision $P_\alpha$ and β-Recall $R_\beta$ are two high-order metrics that measure the overall fidelity and diversity of synthetic data (Ahmed M. Alaa, 2022). $\tilde{X}_g$ and $\tilde{X}_r$ represents the synthetic and real sample, $S_r^\alpha$ and $S_g^\beta$ denotes the α- support of the real distribution and β- support of the synthetic data distribution, respectively. Both metrics range from 0 to 1, where the closer the values are to 1, indicates higher fidelity, diversity and better coverage, calculated by:

$$P_\alpha = \mathbb{P}(\tilde{X}_g \in S_r^\alpha) \quad (16)$$

$$R_\beta = \mathbb{P}(\tilde{X}_r \in S_g^\beta) \quad (17)$$

**Pair-wise Correlation Difference (PCD).** $S_{A,B}$ and $R_{A,B}$ denote the correlation coefficient between column A and B on the real and synthetic data, $R$ and $S$, respectively. Then average the absolute difference for normalization to make sure it falls in the range of [0,1], defined as Equation **错误!未找到引用源。**:

$$PCD(S_{A,B}, R_{A,B}) = \frac{|S_{A,B} - R_{A,B}|}{2} \quad (18)$$

*4.4.3 Crash Frequency Modeling Accuracy Metrics*

Root Mean Squared Error (RMSE) and Mean Squared Error (MSE) are two commonly used metrics to evaluate regression accuracy. MSE represents the average of the squared differences between predicted and observed values, calculated as Equation, where $y_i$ are the observed values, $\hat{y}_i$ are the predicted values, and $n$ is the number of observations. RMSE measures the average magnitude of the errors between predicted and observed values. It's the square root of MSE, calculated as Equation.

$$MSE = \frac{1}{n}\sum_{i=1}^{n}(y_i - \hat{y}_i)^2 \quad (19)$$

$$RMSE = \sqrt{MSE} = \sqrt{\frac{1}{n}\sum_{i=1}^{n}(y_i - \hat{y}_i)^2} \quad (20)$$

4.5 Prediction Interpretation

SHAP (SHapley Additive exPlanations) provides a unified approach to interpreting machine learning model predictions by assigning each feature an importance value based on its contribution to the prediction. This methodology is grounded in the concept of Shapley values from cooperative game theory, ensuring a fair distribution of the "payout" (model prediction) among the features. The SHAP value of feature $i$ is denoted as $\phi_i$, given by:

$$\phi_i = \sum_{S \subseteq N \setminus \{i\}} \frac{|S|!(|N|-|S|-1)!}{|N|!}[f(S \cup \{i\}) - f(S)] \quad (21)$$

where $S \subseteq N$, $N$ is the total set of all features, and $S$ is any subset of $N$ that excluded the feature $i$. $|N|$ and $|S|$ are the number of features in both sets, respectively. $f(S)$ is the model's prediction using the features in subset $S$ and $f(S \cup \{i\})$ is the prediction when feature $i$ is added to subset $S$.

# 5 RESULT ANALYSIS

5.1 Synthetic Data Evaluation

The quality of the synthetic data is evaluated by the similarity between the real and synthetic data, including the coverage similarity and the structural similarity. Coverage similarity is how similar are the distributions of two or more variables in both datasets. Structural similarity is how similar are the underlying patterns and correlations between the two datasets (Vallevik et al., 2024). We also use a metric called $\beta$-recall to access the diversity of the synthetic data.

*5.1.1 Coverage Similarity*

**Conditional density distribution.** 错误！未找到引用源。(1)-(4) illustrates the distribution of real crash data and synthetic data by VAE-Difussion, STaSy, CTGAN, GReaT, and TVAE. As shown in 错误!未找到引用源。(1)-(4), deviations in the distribution of synthetic data based on the proposed generative model—VAE-Difussion—are much smaller than those based on STaSy, CTGAN, GReaT, and TVAE for all variables. For instance, for count variable 'Total Crashes', the distribution generated by TVAE and GReaT show abnormal concentration at certain values. The distribution for the ordinal variable 'Hour' generated by STaSy and TVAE, shows a noticeable shift in the peak values compared to the real data. What's more, the generated distributions of nominal variables are more accurate with VAE-Difussion, compared to the others.

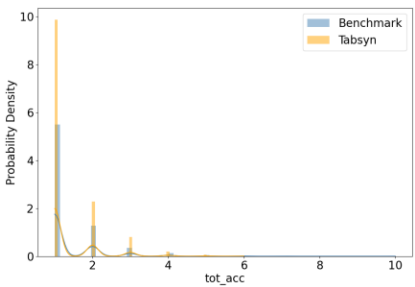 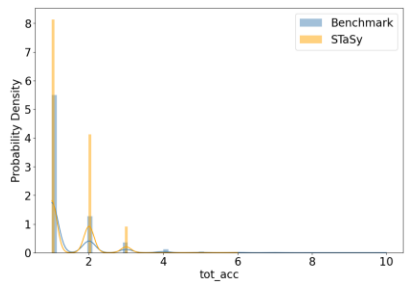 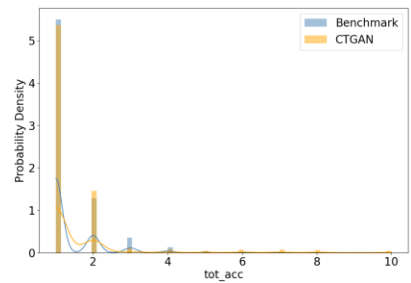

(a) VAE-Difussion  (b) STaSy  (c) CTGAN

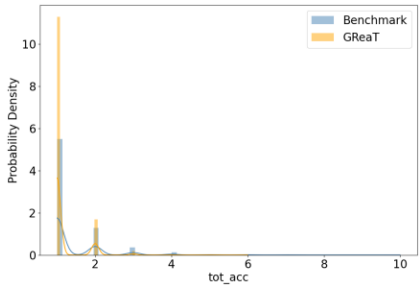

(d) GReaT

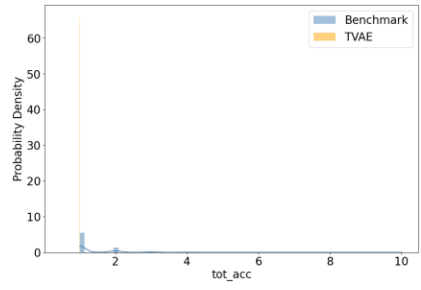

(e) TVAE

(1) Count Variable: Total Crashes

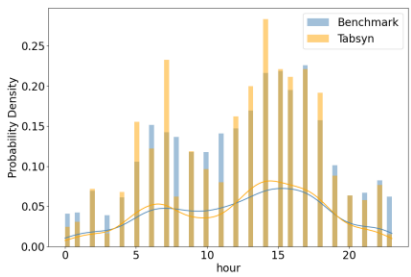

(a) VAE-Difussion

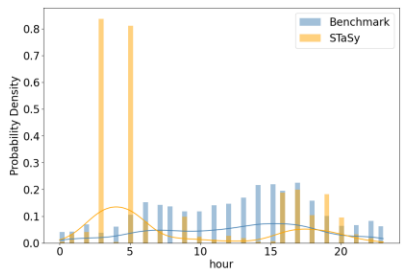

(b) STaSy

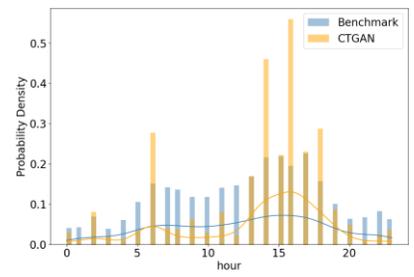

(c) CTGAN

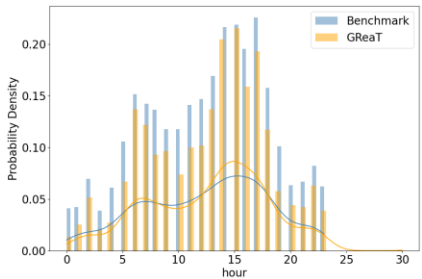

(d) GReaT

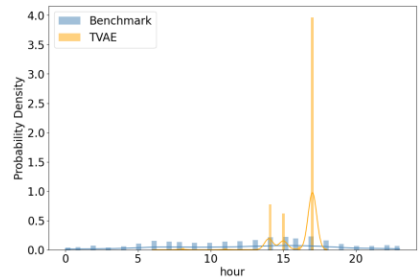

(e) TVAE

(2) Ordinal Variable: Hour

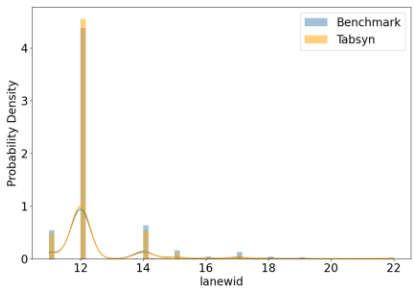

(a) VAE-Difussion

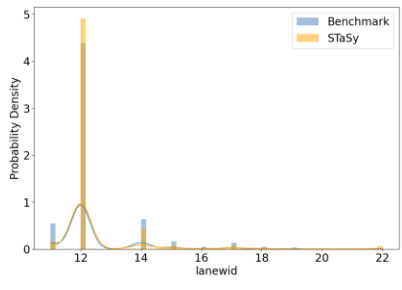

(b) STaSy

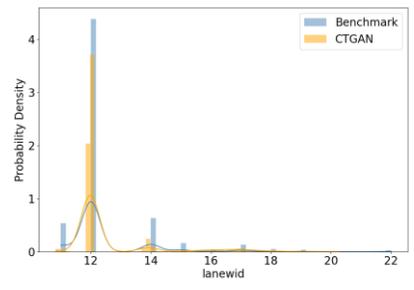

(c) CTGAN

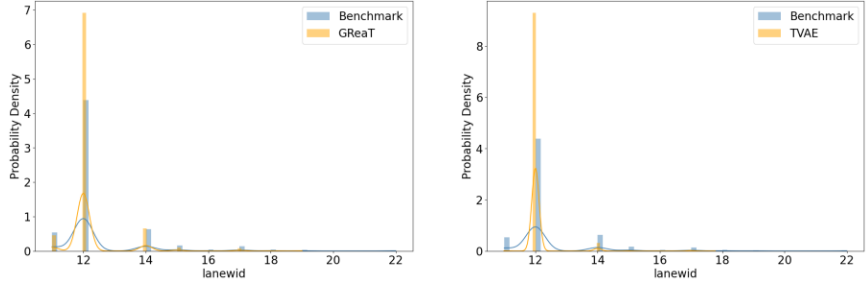

(d) GReaT        (e) TVAE
(3) Real-Valued Variable: Lane Width

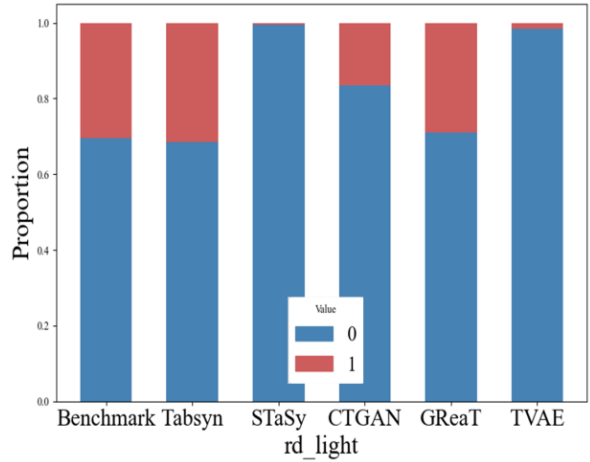

(4) Nominal Variable: Presence of Light

Fig. 2 Conditional Distribution of Synthetic Data

**Joint probability density distribution.** Fig. **3**(1)-(3) display the joint distribution of between variables Total Crashes and Presence of Light; Total Crashes and Hour; as well as Lane Width and Segment Length, respectively. As shown in Fig. **3**(1)-(3), deviations in the synthetic data distribution generated by VAE-Difussion are much smaller than those based on STaSy, CTGAN, GReaT, and TVAE for all variables. The distribution with VAE-Difussion shows two prominent dense areas in Fig. **3**(1) that align well with the real data; while StaSy has significant discrepancies, with only one dense area concentrated in a few narrow bands, indicating it failed to capture the data pattern. Similarly, Fig. **3**(2) shows that VAE-Difussion outperforms the other methods, which exhibit various degrees of deviation from the benchmark data, with STaSy and TVAE showing the most significant discrepancies. As for Fig. **3**(3), GReaT replicates the main dense area, but with some variations in shape and less concentration; TVAE generated distribution shows several discrete points rather than a continuous distribution; CTGAN generated distribution has some additional noise and less concentration; and StaSy generated distribution has a totally different dense area from the real data.

All in all, synthetic data generated by VAE-Difussion has the most similar distribution across all types of variables—count, ordinal, nominal, and real-valued data—justifying its capability in accurately extract the feature patterns of the data.

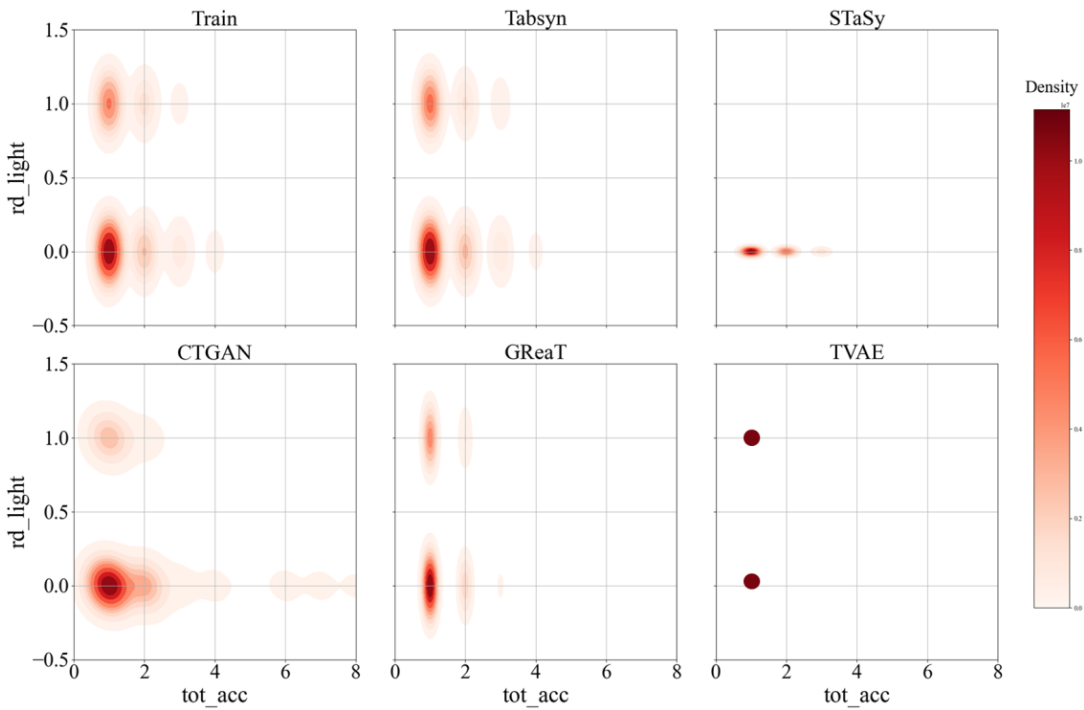

(1) Total Crashes and Presence of Light

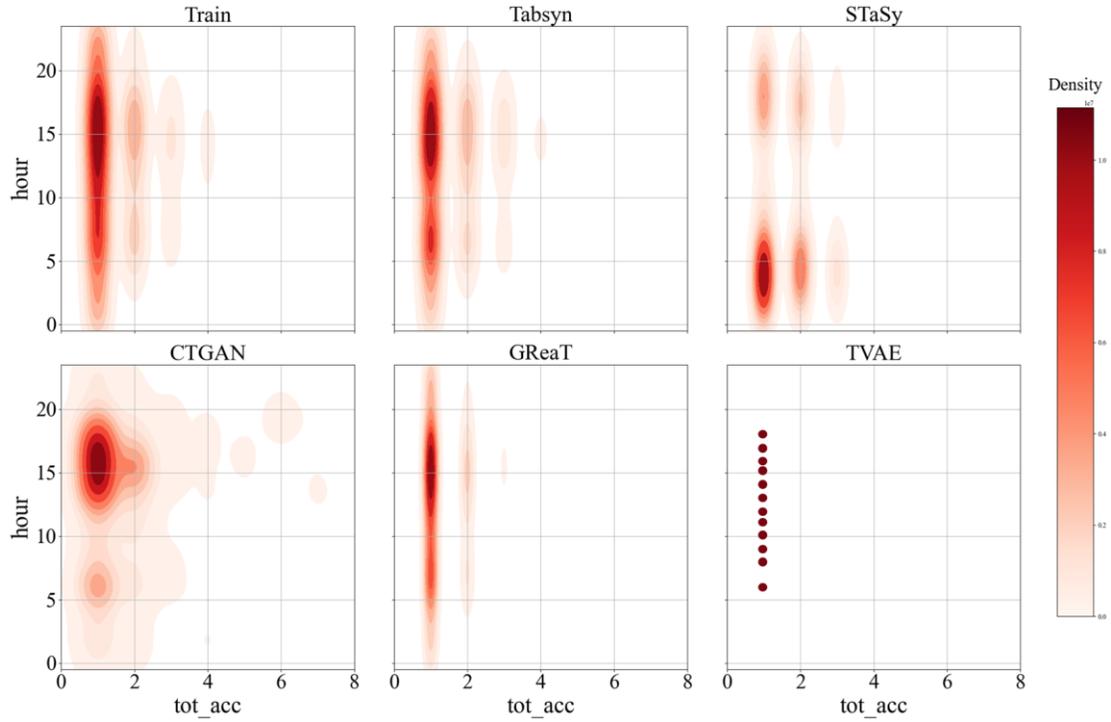

(2) Total Crashes and Hour

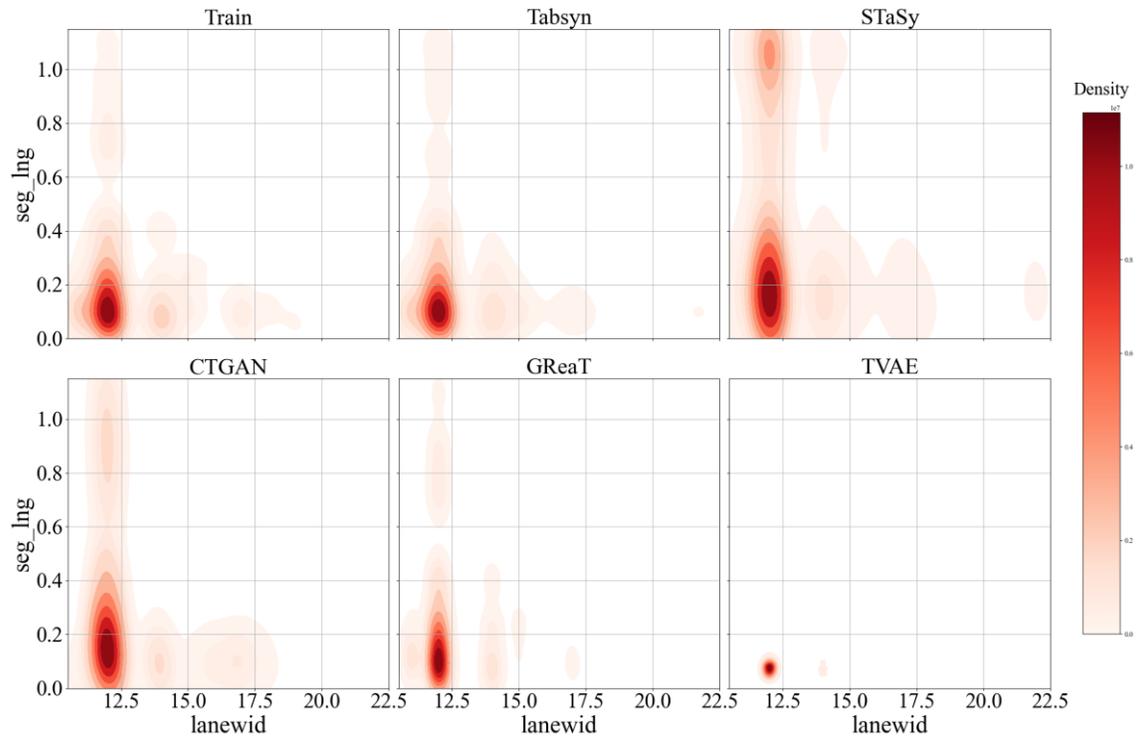

(3) Lane Width and Segment Length

Fig. 3 Joint Probability Distribution of Synthetic Data

In Table 3, we present the results of Detection Score (C2ST), $\alpha$-Precision, and $\beta$-Recall across all methods. For each of these metrics, higher scores indicate better performance.

C2ST measures how difficult it is to distinguish real data from synthetic data, with higher scores indicating better performance. VAE-Difussion scores the highest (0.9745) C2ST, indicating it produces synthetic data very similar to real data. $\alpha$-Precision measures the fraction of synthetic samples that resemble the most typical real samples, where VAE-Difussion also score the highest (0.9825). $\beta$-Recall measures the coverage of real samples by synthetic data, and VAE-Difussion scores the highest again (0.4838), demonstrating good coverage. Overall, VAE-Difussion consistently outperforms other methods across all metrics, closely matching real data in terms of fidelity and diversity.

Table 3 Detection score (C2ST) using logistic regression classifier, α-Precision, and β-Recall

|  | VAE-Difussion | STaSy | CTGAN | GReaT | TVAE |
|---|---|---|---|---|---|
| **C2ST** | **0.9745** | 0.2327 | 0.4396 | 0.7839 | 0.1428 |
| **α-Precision** | **0.9825** | 0.8442 | 0.9648 | 0.5563 | 0.4292 |
| **β-Recall** | **0.4838** | 0.1864 | 0.2282 | 0.0000 | 0.0638 |

*5.1.2 Structural Similarity*

Pair-wise Correlation Difference (PCD). We plot the PCD heatmap for each pair of columns in our data, as shown in Fig. 4. In the plot, lighter colors indicate smaller PCD values between the real and synthetic data, reflecting more accurate correlation estimations. VAE-Difussion gives the most accurate correlation estimation, while other methods exhibit suboptimal performance.

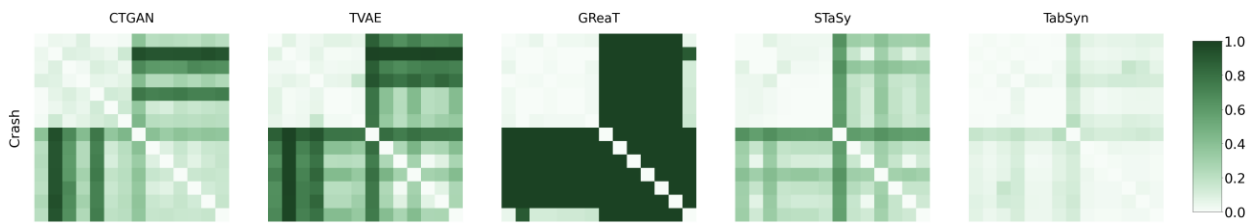

Fig. 4 Pair-wise Correlation Difference of synthetic data and the real data

## 5.2 Crash Frequency Modeling Prediction Accuracy

We employed various generative methods to oversample the original imbalanced dataset. Subsequently, we compared the accuracy of the fitted crash frequency model using the datasets balanced by different methods, alongside the original imbalanced training set. XGBoost was used as the prediction model, with hyperparameters optimized using grid search. The results are presented in Table 4. Smaller RMSE and MSE values indicate better performance. VAE-Difussion demonstrates superior performance with the lowest MSE (0.248) and RMSE (0.498) for overall predictions, outperforming all other methods. For non-zero crash predictions, VAE-Difussion again excels with the lowest MSE (0.720) and RMSE (0.849), significantly better than the original training set and other models.

Table 4 Prediction Accuracy of Crash Frequency Models

|  | Original training set | VAE-Difussion | CTGAN | TVAE | GReaT | STaSy | ZIP |
|---|---|---|---|---|---|---|---|
| *Overall Performance* | | | | | | | |
| MSE | 0.285 | **0.248** | 0.298 | 0.281 | 0.392 | 0.257 | 0.273 |
| RMSE | 0.534 | **0.498** | 0.546 | 0.530 | 0.626 | 0.508 | 0.521 |
| | | | | | | | |
| *Non-Zero Prediction* | | | | | | | |
| MSE | 1.607 | **0.720** | 1.397 | 1.454 | 0.892 | 0.866 | 1.600 |
| RMSE | 1.268 | **0.849** | 1.182 | 1.206 | 0.944 | 0.930 | 1.230 |

## 5.3 Comparison with Statistical Crash Frequency Model

We compare our proposed XGBoost model with the traditional statistical model developed to solve the excessive zero observations—Zero-inflated Poisson (ZIP)—to demonstrate the effectiveness of the proposed machine learning model in accurately predicting crash frequency.

The prediction accuracy is quantified using metrics, MSE and RMSE, for evaluating both overall and non-zero crash predictions, as shown in Table 4. Our proposed method—VAE-Difussion—has the best performance across all metrics and outperforms the traditional statistical model—Zero-inflated Poisson (ZIP). Additionally, all data-driven models (VAE-Difussion, CTGAN, TVAE, GReaT, and STaSy) show better performance compared to the ZIP, indicating the superiority of advanced machine learning techniques over traditional statistical approaches in predicting crash frequency.

5.4 Discussion on Key Features

*5.4.1 SHAP Post-hoc Analysis*

The SHAP summary plot (see Fig. 5) illustrates the impact and direction of various input features on the predicted crash count for a given road segment. The colors represent the magnitude of each feature, with blue points indicating lower feature values and red points representing higher values. The SHAP values on the x-axis show the features' positive or negative influence on the model's output, where a positive SHAP value suggests the model predicts a higher crash count and a negative SHAP value indicates a lower predicted crash count.

Fig. 5 also ranks the importance of these features from top to bottom, where AAHT is the most influential feature, followed by Segment Length, Hour, Grade Percentage, etc. AAHT (Annual Average Hourly Traffic) has the greatest impact on the predicted crash count, with higher AAHT values (in red) contributing to an increase in crashes. The result is consistent with previous studies (Hou et al., 2021; Hoye and Hesjevoll, 2020). This phenomenon is likely explained by the fact that increased traffic volume leads to more vehicle conflicts, while high vehicle density and congestion may reduce available reaction times and increase the likelihood of crashes. Additionally, prolonged congestion or stop-and-go traffic causes driver fatigue, which further elevates crash risk. Segment Length ranks second in importance. Longer road segments (in red) tend to show more crashes simply because they encompass a larger area, leading to a higher overall crash count due to the greater length being analyzed. The Hour variable is also highly influential. The Hour variable is also highly influential, with rush hour periods (in red) linked to a higher crash count. During these hours, increased traffic flow, time pressure, and driver fatigue raise the likelihood of accidents. Additionally, road geometry features like Grade Percentage, median width, surface type, and curvature degree also significantly impact crash count.

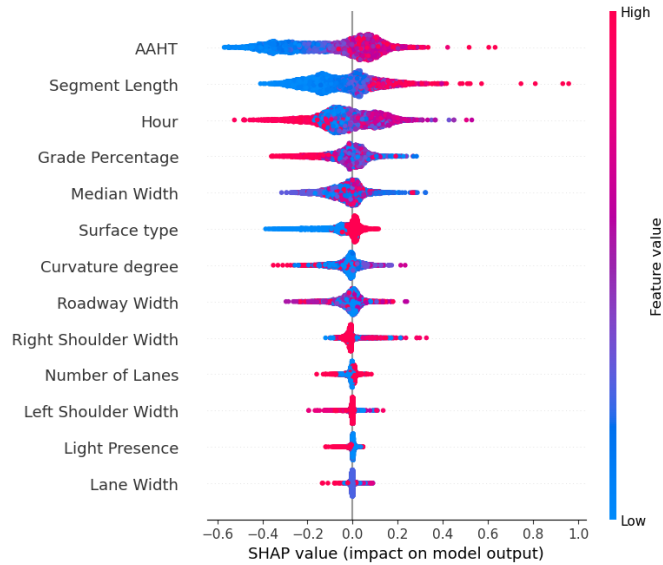

Fig.5 SHAP Value Summary Plot

The SHAP summary plot gives an overview of feature importance, while the SHAP dependence plot in Fig. 8 shows how the model's predictions change with different feature values. Each dot represents a road segment, and the vertical spread at a single feature value reflects the interaction effects between features.

In Fig.8 (a), the x-axis represents road surface type, where $x=0$ corresponds to asphalt and $x=1$ to concrete. The SHAP values for asphalt roads ($x=0$) are mostly negative, indicating that asphalt pavement may help reduce the predicted crash count and improve safety. In contrast, the SHAP values for concrete roads ($x=1$) are higher, showing that concrete surfaces could increase the predicted crash count. One potential explanation is that worn concrete surfaces can become slick, particularly in adverse weather conditions such as rain or snow, which diminishes skid resistance and elevates the risk of crashes. Furthermore, roads with concrete pavement often undergo less frequent repair and maintenance compared to those with asphalt pavement, which may result in surface deterioration over time and contribute to an increased accident rate.

Fig.8 (b) shows the effect of lighting on the predicted crash count. The x-axis represents light presence (0: light present; 1: no light). Under lit conditions ($x=1$), some data points have negative SHAP values, which indicates that the presence of light reduces the predicted crash count, suggesting that lighting helps lower crash risk to some extent. Research shows that accidents are more likely in darkness, particularly on unlit roads. Road lighting can reduce nighttime accidents by about 50%, with a slightly greater effect on fatal accidents (Wanvik, 2009a).

This analysis suggests several practical safety recommendations. Regular inspections and maintenance should be carried out on concrete roads to prevent deterioration. On road sections with concrete pavement, stricter safety measures, like lower speed limits or barriers, can help reduce crash count. For roads without sufficient lighting, installing more streetlights or reflective markers can improve safety, especially in low-visibility conditions and at night.

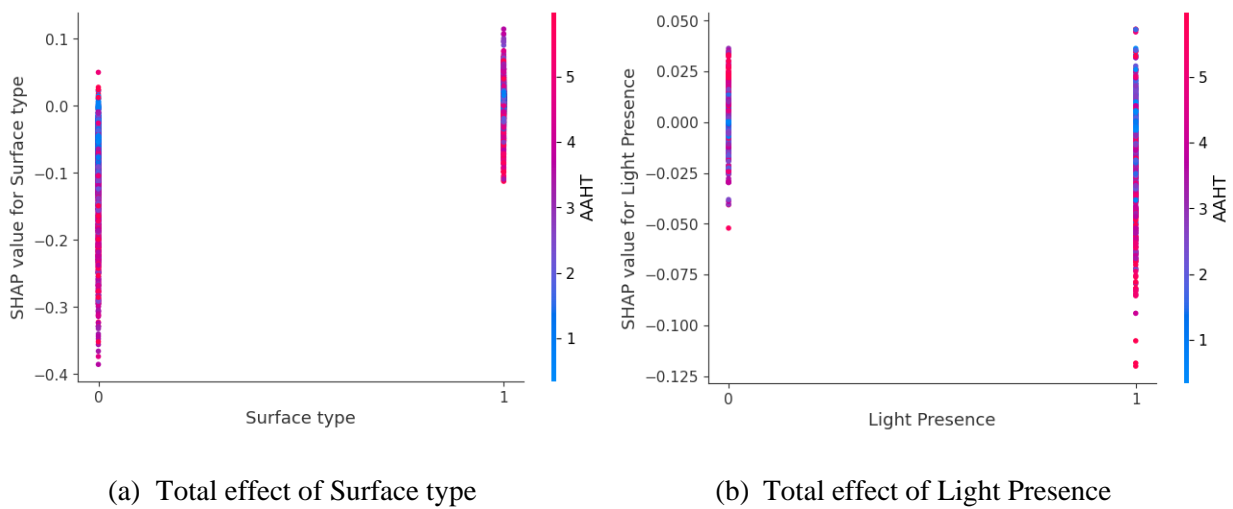

(a) Total effect of Surface type  (b) Total effect of Light Presence

Fig. 6 SHAP total effect plots

The dependence plot for SHAP interaction values helps distinguish main effects from interaction effects. Fig. 9 illustrates the main effects of variables like Hour, AAHT, Lane Width, and Grade Percentage, along with selected interaction effects. Main effects capture the standalone contribution of each feature, while interaction effects reveal how pairs of features together influence the prediction.

Fig. 9 (a) shows that the SHAP main effect values between 00:00 and 18:00 are mostly positive when peak crash periods occur between 13:00 and 17:00. After 18:00, the values gradually decrease, reaching their lowest point at 21:00, before rising again. Figure 9(b) indicates that road segments with asphalt pavement significantly reduce predicted crash count during the peak accident period (13:00 to 17:00). One possible explanation for this pattern is the increased traffic volume and driver fatigue during the afternoon peak hours, which contributes to higher crash risk. The reduced crash count on asphalt pavements could be due to better traction and surface conditions, as well as clearer road markings and signage on asphalt surfaces, making it safer compared to concrete during these high-risk periods.

Therefore, safety controls should be enhanced during peak accident hours (13:00–17:00) by increasing traffic monitoring and reducing accident response times. This will help minimize delays and lower the risk of secondary collisions. Asphalt surfaces, which improve vehicle handling and safety, should be prioritized on high-risk road segments to reduce accident rates.

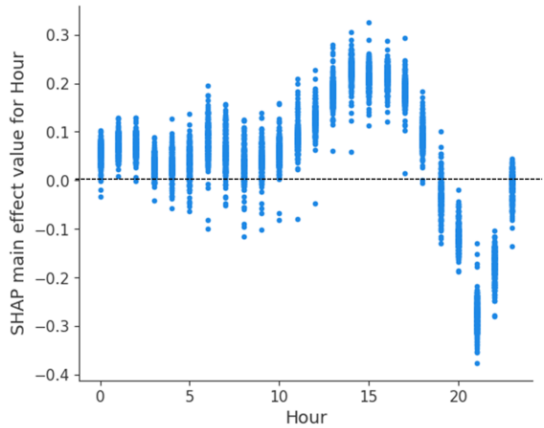 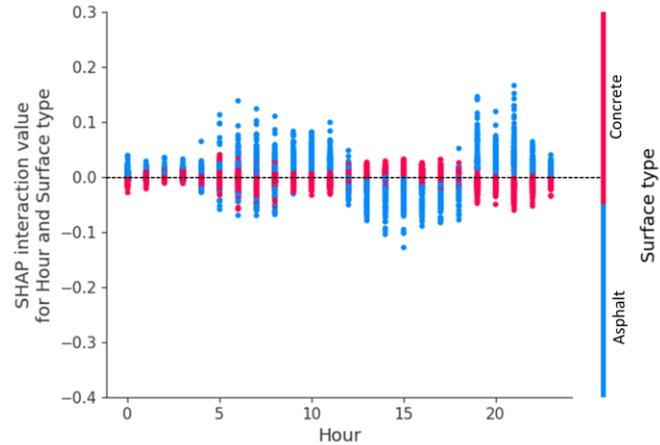

(a) Main effect of Hour    (b) Interaction effect of Hour

In Fig. 9(c), the main effect of AAHT shows that at low levels of traffic (AAHT between 0 and 2), the accident risk is relatively low. As AAHT increases to a moderate level (2 to 4), the accident risk begins to rise, and at high AAHT levels (above 4), the risk increases sharply. This indicates that higher traffic volumes are associated with a significant increase in predicted crash counts. Fig. 9(d) examines the interaction effect of AAHT and light presence. Under high AAHT conditions, the SHAP interaction values with light presence (in red) are close to or below zero. This suggests that, in high-traffic scenarios, the presence of lighting may reduce accident risk in certain cases. This finding can be attributed to the role of lighting in improving visibility and overall road safety, especially under high traffic conditions. Adequate lighting enhances the visual field, allowing drivers to detect other vehicles more easily. Additionally, lighting helps drivers adapt more quickly to changing light conditions, maintain better alertness, and adhere to traffic rules, which is crucial during periods of high traffic volume. It also aids in minimizing speed-related accidents by providing a clearer view of the road, contributing to overall safer driving behavior in high-traffic scenarios (Wanvik, 2009b).

To improve road safety, it may be beneficial to install and optimize street lighting on high-traffic road segments and key intersections. Additionally, in areas with insufficient lighting, using supplementary measures such as reflective signs and warning lights might further reduce accident risks, especially under poor weather or low-visibility conditions.

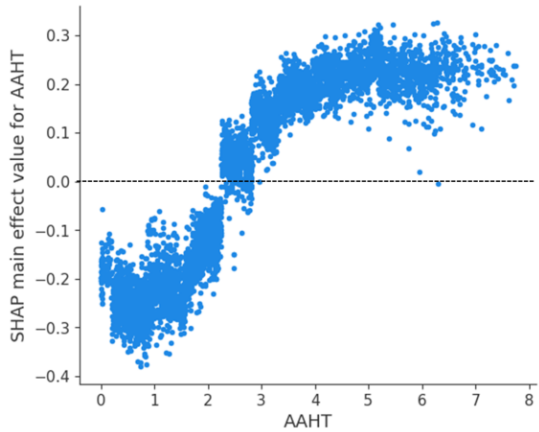 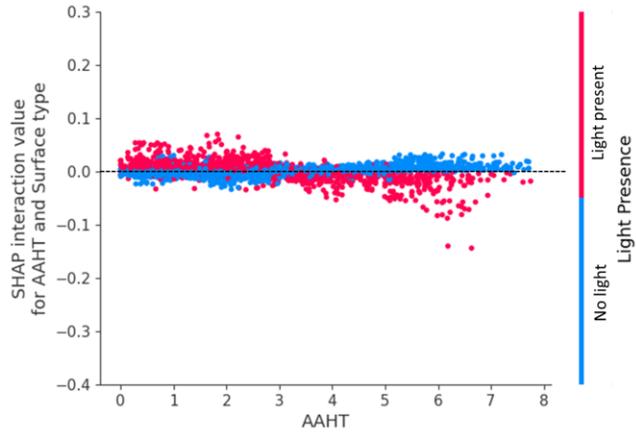

(c) Main effect of AAHT  (d) Interaction effect of AAHT

Fig. 9(e) shows that as Lane Width increases, the main effect value moves closer to zero, indicating that wider lanes reduce accident risk. Since wider lanes offer drivers more space and maneuverability, lowers the chance of vehicle collisions and improves road safety. In Figure 9(f), narrower lanes exhibit a notable decrease in accident risk when matched with a smaller Curvature Degree. In contrast, wider lanes mitigate the effects of higher curvature, reducing the potential risk of accidents caused by sharp curves. This suggests that wider lanes offset the impact of curvature on accident risk, helping to reduce the likelihood of crashes.

To improve road safety, increasing lane width on high-risk segments, especially on roads with higher curvature, reduces accident risk. Implementing speed limits and adding clear signage on narrow lanes with low curvature ensures safer driving conditions and decreases the likelihood of accidents.

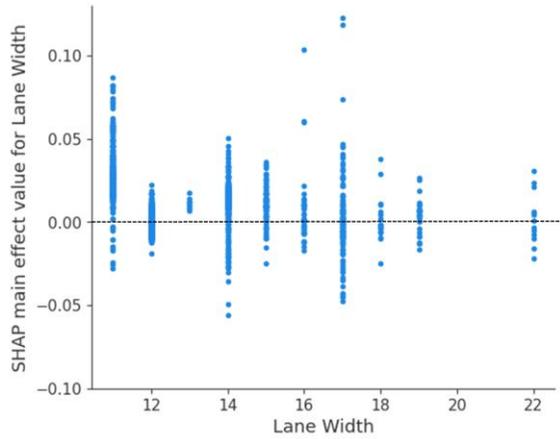
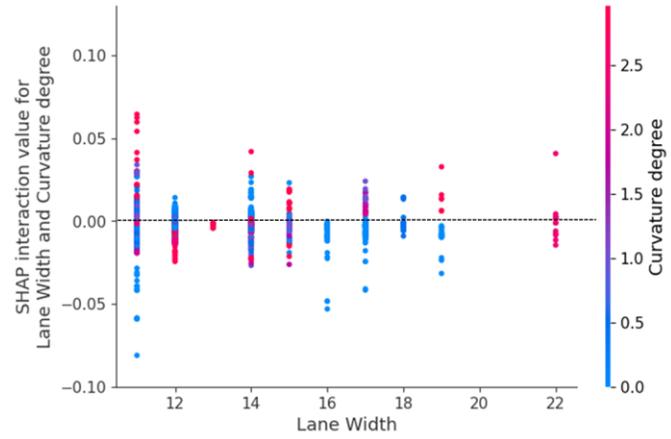

(e) Main effect of Lane Width  (f) Interaction effect of Lane Width

Fig. 9(g) shows that when the Grade Percentage is positive (indicating a higher proportion of uphill segments), SHAP values decrease significantly, especially when the Grade Percentage exceeds 3%. This suggests that road sections with more uphill segments experience a lower accident risk, likely due to vehicles moving more slowly as they require more power to ascend. Conversely, when the Grade Percentage is negative (indicating a higher proportion of downhill segments), SHAP values increase, indicating a higher accident risk. This could be because vehicles tend to accelerate more easily on downhill sections, increasing the likelihood of losing control. Fig. 9(h) demonstrates that on road sections with a high proportion of uphill segments (Grade Percentage > 3%), wider left shoulders are associated with negative interaction values, further reducing accident risk. The additional space provided by the wider left shoulder likely offers drivers more room for maneuvering or stopping safely in emergencies, thereby lowering the chances of accidents.

On roads with a high proportion of uphill and downhill segments, increasing shoulder width can enhance safety. A wider shoulder provides a larger buffer zone, allowing drivers more room to maneuver, especially during emergencies. This additional space helps to reduce accident risks by offering better control and response options for drivers.

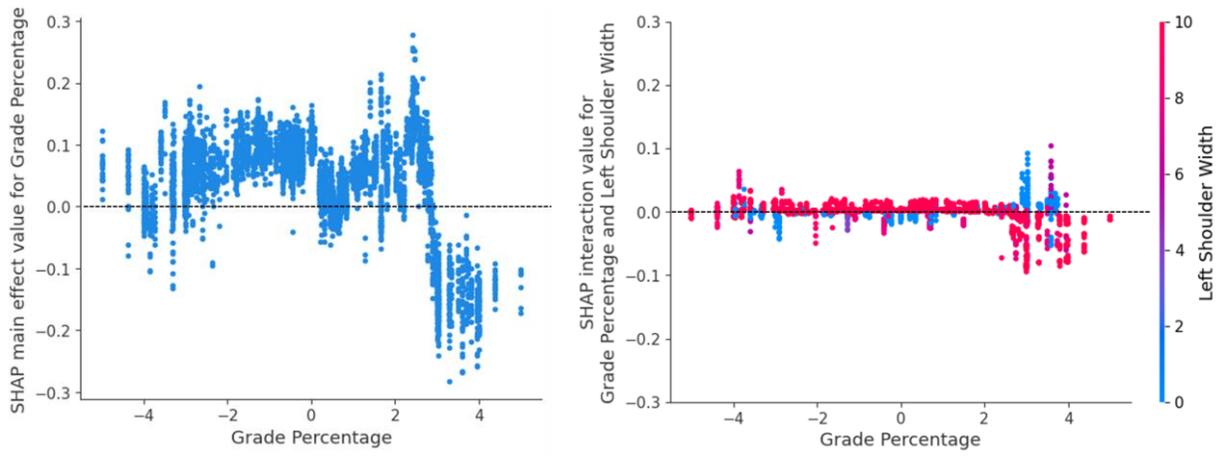

(g) Main effect of Grade Percentage    (h) Interaction effect of Grade Percentage

Fig. 7 SHAP main and interaction effects plots

*5.4.2 Policy Recommendation*

To enhance road safety, targeted improvements to infrastructure and traffic management strategies should be prioritized. For road infrastructure enhancements, upgrading high-risk road segments with durable materials like asphalt and implementing proactive maintenance schedules for concrete surfaces can improve traction and prevent surface deterioration, particularly in adverse weather conditions. Expanding road lighting coverage at intersections and curves is essential for better nighttime visibility, while widening lanes and shoulders on steep gradients and sharp curves can provide additional maneuvering space to reduce crash risks. In areas where physical expansion is impractical, advanced reflective markers and dynamic signage can serve as effective alternatives. Traffic management measures, such as adaptive monitoring systems and variable speed limits during peak hours, can help alleviate congestion and lower crash risks. Stricter safety interventions, including reduced speed limits and protective barriers, should be applied in locations with high crash frequencies. These measures, combined with public education campaigns, aim to create a safer and more resilient transportation network.

## 6 CONCLUSIONS

In this study, to address excessive zero observations and mitigate the limitations of traditional crash data collection methods, we develop a hybrid VAE-Diffusion deep generative model to generate realistic crash data. This method considers the multi-type features of the tabular crash data, i.e. count, ordinal, nominal, and real-valued. Through a comprehensive evaluation of the quality of synthetic data using various metrics, the proposed VAE-Difussion is verified to have the best generation performance. The synthetic data generated by VAE-Difussion demonstrates the best quality in terms of similarity, accuracy, diversity, and structural consistency. We compare the prediction accuracy with data generated by different deep generative methods, results show that VAE-Difussion has the lowest MSE and RMSE (0.248 and 0.498 for overall predictions; 0.720 and 0.849 for non-zero crash predictions) and surpasses all other methods. In addition, all data-driven models (VAE-Difussion, CTGAN, TVAE, GReaT, and STaSy) show better prediction accuracy compared to the statistical Zero-inflated Poisson (ZIP) model, indicating the proposed XGBoost model provides a higher level of prediction accuracy. We also employ SHAP to interpret crash frequency prediction results, identify key explanatory factors influencing crash counts, and provide corresponding safety policy recommendations to aid governmental agencies in targeting high-risk areas and optimizing resource allocation.

In future work, we will investigate the effectiveness of the proposed method in balancing crash data that includes omitted variables, heteroscedasticity, and endogeneity issues. What's more, we intend to examine how resampling ratios of the majority zero-crash records to the minority non-zero records impact the prediction accuracy and probability distribution. Furthermore, although this study explored the variables correlation between crash frequency and road factors (e.g., road geometry), traffic volume, and hour, etc. with the proposed VAE-Diffusion deep generative model, future research can delve into the causal effects of these factors on crash frequency.